\documentclass[10pt,twocolumn,letterpaper]{article}
\pdfoutput=1
\usepackage[pagenumbers]{cvpr} 

\usepackage{graphicx}
\usepackage{amsmath}
\usepackage{amssymb}
\usepackage{booktabs}

\usepackage{enumitem}
\usepackage{multirow}

\usepackage[pagebackref,breaklinks,colorlinks]{hyperref}

\usepackage[capitalize]{cleveref}
\crefname{section}{Sec.}{Secs.}
\Crefname{section}{Section}{Sections}
\Crefname{table}{Table}{Tables}
\crefname{table}{Tab.}{Tabs.}

\def\cvprPaperID{1349} 
\def\confName{CVPR}
\def\confYear{2023}

\begin{document}

\title{Graphics Capsule: Learning Hierarchical 3D Face Representations \\ from 2D Images}

\author{Chang Yu$^{1,2}$, Xiangyu Zhu$^{1,2}$\footnotemark[1], Xiaomei Zhang$^{1,2}$, Zhaoxiang Zhang$^{1,2,3}$, Zhen Lei$^{1,2,3}$\\
    $^{1}$State Key Laboratory of Multimodal Artificial Intelligence Systems,\\ Institute of Automation, Chinese Academy of Sciences\\
    $^{2}$School of Artificial Intelligence, University of Chinese Academy of Sciences\\
    $^{3}$ Centre for Artificial Intelligence and Robotics, Hong Kong Institute of Science \& Innovation,\\ Chinese Academy of Sciences\\
    {\tt\small \{chang.yu, xiangyu.zhu, zlei\}@nlpr.ia.ac.cn}\\
    {\tt\small \{zhangxiaomei2016, zhaoxiang.zhang\}@ia.ac.cn}\\
}
\maketitle

{\renewcommand{\thefootnote}{\fnsymbol{footnote}}
    \footnotetext[1]{Corresponding author.}}

\begin{abstract}
The function of constructing the hierarchy of objects is important to the visual process of the human brain. Previous studies have successfully adopted capsule networks to decompose the digits and faces into parts in an unsupervised manner to investigate the similar perception mechanism of neural networks. However, their descriptions are restricted to the 2D space, limiting their capacities to imitate the intrinsic 3D perception ability of humans. In this paper, we propose an Inverse Graphics Capsule Network (IGC-Net) to learn the hierarchical 3D face representations from large-scale unlabeled images. The core of IGC-Net is a new type of capsule, named graphics capsule, which represents 3D primitives with interpretable parameters in computer graphics (CG), including depth, albedo, and 3D pose. Specifically, IGC-Net first decomposes the objects into a set of semantic-consistent part-level descriptions and then assembles them into object-level descriptions to build the hierarchy. The learned graphics capsules reveal how the neural networks, oriented at visual perception, understand faces as a hierarchy of 3D models. Besides, the discovered parts can be deployed to the unsupervised face segmentation task to evaluate the semantic consistency of our method. Moreover, the part-level descriptions with explicit physical meanings provide insight into the face analysis that originally runs in a black box, such as the importance of shape and texture for face recognition. Experiments on CelebA, BP4D, and Multi-PIE demonstrate the characteristics of our IGC-Net.
\end{abstract}

\section{Introduction}
\label{sec:intro}
A path toward autonomous machine intelligence is to enable machines to have human-like perception and learning abilities~\cite{lecun2022path}. As humans, by only observing the objects, we can easily decompose them into a set of part-level components and construct their hierarchy even though we have never seen these objects before. This phenomenon is supported by the psychological studies that the visual process of the human brain is related to the construction of the hierarchical structural descriptions~\cite{marr2010vision,hinton1979some,marr1978representation,singh2001part}. To investigate the similar perception mechanism of neural networks, previous studies~\cite{kosiorek2019stacked,yu2022hp} incorporate the capsule networks, which are designed to present the hierarchy of objects and describe each entity with interpretable parameters. After observing a large-scale of unlabeled images, these methods successfully decompose the digits or faces into a set of parts, which provide insight into how the neural networks understand the objects. However, their representations are limited in the 2D space. Specifically, these methods follow the analysis-by-synthesis strategy in model training and try to reconstruct the image by the decomposed parts. Since the parts are represented by 2D templates, the reconstruction becomes estimating the affine transformations to warp the templates and put them in the right places, which is just like painting with stickers. This strategy performs well when the objects are intrinsically 2D, like handwritten digits and frontal faces, but has difficulty in interpreting 3D objects in the real world, especially when we want a view-independent representation like humans~\cite{biederman1987recognition}.

\begin{figure*}[!ht]		
	\begin{center}                                              \includegraphics[width=1\linewidth]{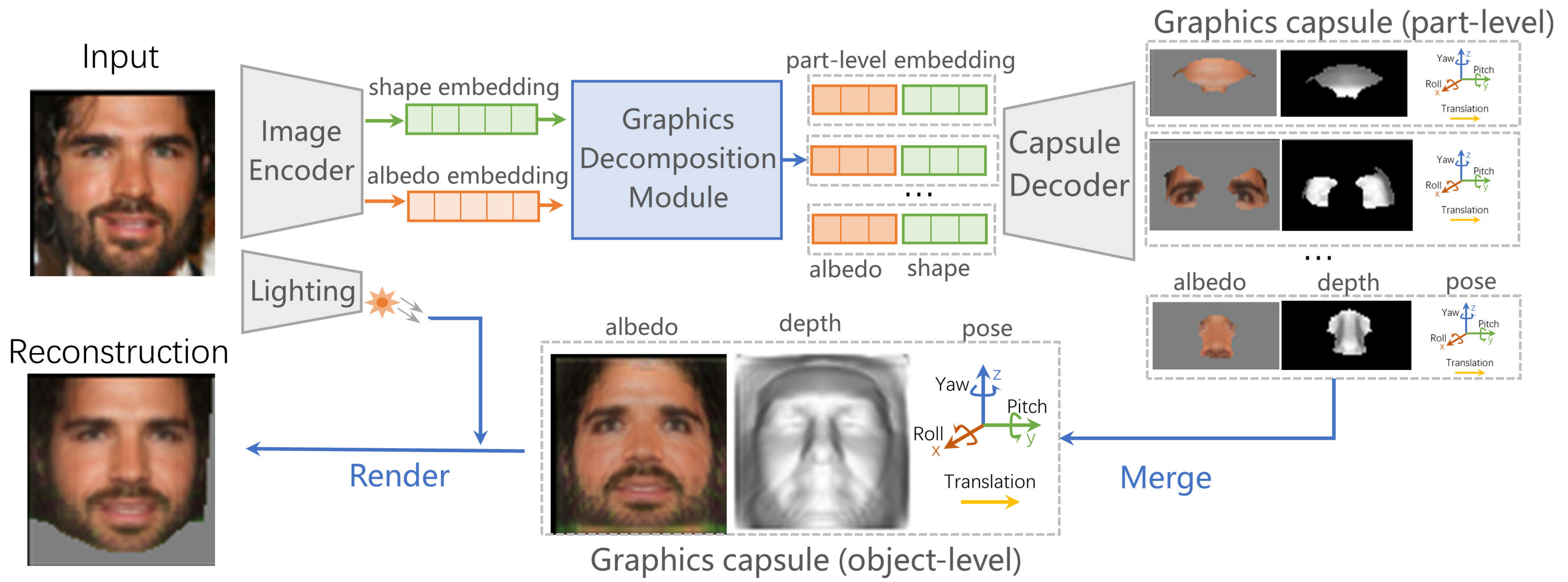}
	\end{center}
	\vspace{-10px}
	\caption{Overview of the Inverse Graphics Capsule Network (IGC-Net). The input image is first encoded to global shape and albedo embeddings and then sent to a decomposition module to get the spatially-decoupled part-level graphics capsules. Afterward, these capsules are decoded to get the explicit 3D descriptions of parts, which are assembled by their depth to generate the object capsules as object-centered representations. Finally, the object capsules are illuminated, posed, and rendered to fit the input image. After training, the learned graphics capsules naturally build hierarchical 3D representations.}
	\label{fig-overview}
\end{figure*}

How to represent the perceived objects is the core research topic in computer vision~\cite{poggio1990network,biederman1993recognizing}. One of the most popular theories is the Marr's theory~\cite{marr1978representation,marr2010vision}. He believed that the purpose of the vision is to build the descriptions of shapes and positions of things from the images and construct hierarchical 3D representations of objects for recognition. In this paper, we try to materialize Marr's theory on human faces and propose an Inverse Graphics Capsule Network (IGC-Net), whose primitive is a new type of capsule (i.e., graphics capsule) that is defined by computer graphics (CG), to learn the hierarchical 3D representations from large-scale unlabeled images. Figure~\ref{fig-overview} shows an overview of the proposed method. Specifically, the hierarchy of the objects is described with the part capsules and the object capsules, where each capsule contains a set of interpretable parameters with explicit physical meanings, including depth, albedo, and pose. During training, the input image is first encoded to a global shape and albedo embeddings, which are sent to a decomposition module to get the spatially-decoupled part-level graphics capsules. Then, these capsules are decoded by a shared capsule decoder to get explicit 3D descriptions of parts. Afterward, the parts are assembled by their depth to generate the object capsules as the object-centered representations, naturally constructing the part-object hierarchy. Finally, the 3D objects embedded in the object capsules are illuminated, posed, and rendered to fit the input image, following the analysis-by-synthesis manner. When an IGC-Net is well trained, the learned graphics capsules naturally build hierarchical 3D representations.

We apply IGC-Net to human faces, which have been widely used to investigate human vision system~\cite{tanaka2016parts} due to the similar topology structures and complicated appearances. Thanks to the capacity of the 3D descriptions, IGC-Net successfully builds the hierarchy of in-the-wild faces that are captured under various illuminations and poses. We evaluate the IGC-Net performance on the unsupervised face segmentation task, where the silhouettes of the discovered parts are regarded as segment maps. We also incorporate the IGC-Net into interpretable face analysis to uncover the mechanism of neural networks when recognizing faces. 

The main contributions of this paper are summarized as:
\begin{itemize}
	\item This paper proposes an Inverse Graphics Capsule Network (IGC-Net) to learn the hierarchical 3D face representations from unlabeled images. The learned graphics capsules in the network provide insight into how the neural networks, oriented at visual perception, understand faces as a hierarchy of 3D models. 
	
	\item A Graphics Decomposition Module (GDM) is proposed for part-level decomposition, which incorporates shape and albedo information as cues to ensure that each part capsule represents a semantically consistent part of objects.
	
	\item We execute the interpretable face analysis based on the part-level 3D descriptions of graphics capsules. Besides, the silhouettes of 3D parts are deployed to the unsupervised face segmentation task. Experiments on CelebA, BP4D, and Multi-PIE show the effectiveness of our method.
	
\end{itemize}
\section{Related Work}
\label{gen_inst}

\subsection{Capsule Network}
The connections of the human brain are thought to be sparse and hierarchical~\cite{barrett2012hierarchical, bodegaard2001hierarchical,jaaskelainen2022sparse,georgopoulos1986neuronal}, which inspires the design of capsule networks to present the objects with dynamic parse trees. Given inputs, capsule networks~\cite{hinton2011transforming,sabour2017dynamic,hinton2018matrix,kosiorek2019stacked,sabour2021unsupervised,yu2022hp} will encode the images to a set of low-level capsules, which describe the local entities of the objects, and then assemble them into higher-level capsules to describe more complicated entities. The parameters of capsules are usually with explicit meanings, which enables the interpretability of neural networks. Recently, some capsule networks have been proposed to explore the hierarchy of objects. SCAE~\cite{kosiorek2019stacked} proposes to describe the objects with a set of visualizable templates through unsupervised learning. However, SCAE can only handle simple 2D objects like digits. HP-Capsule~\cite{yu2022hp} extends SCAE to tackle human faces, which proposes subpart-level capsules and uses the compositions of subparts to present the variance of pose and appearance. Due to the limitation of 2D representations, HP-Capsule can only tackle faces with small poses. Sabour et al.~\cite{sabour2021unsupervised} propose to apply the capsule network to human bodies, but it needs optical flow as additional information to separate the parts. In this paper, we propose graphics capsules to learn the hierarchical 3D representations from unlabeled images.

\subsection{Unsupervised Part Segmentation}
We evaluate the graphics capsule performance on the unsupervised face segmentation task. Several methods have been proposed for this challenging task. DFF~\cite{collins2018deep} proposes to use non-negative matrix factorization upon the CNN features to discover semantics, but it needs to optimize the whole dataset during inference. 
Choudhury et al.~\cite{choudhury2021unsupervised} follow a similar idea, which uses k-means to cluster the features obtained by a pre-trained network. SCOPS~\cite{hung2019scops} and Liu et al.~\cite{liu2021unsupervised} propose to constrain the invariance of images between TPS transformation. However, their methods rely on the concentration loss to separate parts, leading to similar silhouettes of different parts. HP-Capsule~\cite{yu2022hp} proposes a bottom-up schedule to aggregate parts from subparts. The parts of the HP-Capsule rely on the learning of subpart-part relations, which is unstable when tackling faces with large poses. Compared with these methods, our IGC-Net can provide interpretable 3D representations of the parts, which are with salient semantics and keep semantic consistency across the in-the-wild faces with various poses.

\subsection{Unsupervised 3D Face Reconstruction}
Learning to recover the 3D face from 2D monocular images has been studied for years. Following the analysis-by-synthesis strategy, many methods~\cite{tran2018nonlinear,zhou2019dense,chen2020self,deng2019accurate} propose to estimate the parameters of the 3D Morphable Model~\cite{paysan20093d}, which describes the faces with a uniform topology pre-defined by humans. Recently, several works~\cite{wu2020unsupervised,zhang2021learning,zhang2022physically} have been proposed to only use the symmetric character of faces to learn 3D face reconstruction. Under the graphics decomposition, these methods achieve promising results. Inspired by them, we propose the graphics capsule to learn the hierarchical 3D face representations from images, which provides insight into how neural networks understand faces by learning to decompose them into a set of parts.

\section{Method}
\label{headings}
Based on previous explorations in capsule networks~\cite{kosiorek2019stacked, yu2022hp}, our goal is to explore a system that can build hierarchical 3D representations of objects through browsing images. Specifically, we focus on the human faces and aim to learn the part-object hierarchy in an unsupervised manner, where each part is represented by a set of interpretable CG parameters, including shape, albedo, 3D poses, etc. In the following sections, we will introduce the graphics capsule and the overview of the network in Section~\ref{method-overview}, the graphics decomposition module that is used to build hierarchy in Section~\ref{method-attention}, and the loss functions that enable unsupervised learning in Section~\ref{method-loss}.

\subsection{Overview}
\label{method-overview}
To learn a hierarchical 3D representation from unlabeled images, we propose an Inverse Graphics Capsule Network (IGC-Net), whose capsules are composed of interpretable CG descriptions, including a depth map $\mathbf{D}\in\mathbb{R}^{H\times W}$, an albedo map $\mathbf{A}\in\mathbb{R}^{C\times H\times W}$ and 3D pose parameters $\mathbf{p}\in\mathbb{R}^{1\times6}$ (rotation angles and translations). Our IGC-Net is applied to human faces, which have been widely used to investigate the human vision system due to their similar topology structures and complicated appearances. The overview of IGC-Net is shown in Figure~\ref{fig-overview}. Following a bottom-up schedule, a CNN-based image encoder first encodes the input image $\mathbf{I}$ into the shape and the albedo embeddings $\mathbf{f}_s$ and $\mathbf{f}_a$:
\begin{align}\label{equ-image-encoder}
	\mathbf{f}_s, \mathbf{f}_a&=\mathrm{ImageEncoder}(\mathbf{I}).
\end{align}
Then a Graphics Decomposition Module (GDM) is employed to decompose the global embeddings into a set of part-level embeddings, which can be further decoded into interpretable graphics capsules:
\begin{align}\label{equ-overview-gdm}
	\begin{aligned}
		\{\hat{\mathbf{e}}_s^1, ..., \hat{\mathbf{e}}_s^M\},\{ \hat{\mathbf{e}}_a^1, ..., \hat{\mathbf{e}}_a^M\}&=\mathrm{GDM}(\mathbf{f}_s, \mathbf{f}_a),\\
		\{\mathbf{D}_p^m, \mathbf{A}_p^m, \mathbf{p}_p^m\}&=\mathrm{GraphicsDecoder}(\hat{\mathbf{e}}_s^m, \hat{\mathbf{e}}_a^m),\\
	\end{aligned}
\end{align}
where  $\hat{\mathbf{e}}_s^m$ is the shape embedding of the $m$th part, $\hat{\mathbf{e}}_a^m$ is the corresponding albedo embedding, $M$ is the number of part capsules, and $\Theta_p^m:\{\mathbf{D}_p^m, \mathbf{A}_p^m, \mathbf{p}_p^m\}$ is a graphics capsule that describes a part with depth, albedo, and 3D pose. Afterward, the part capsules are assembled according to their depth to generate the global object capsule:
\begin{equation}\label{equ-overview-assemble}
	\begin{aligned}
		\mathbf{V}^m(i,j)&=\mathbf{1}_{m=\mathop{\mathrm{argmin}}\limits_{n}(\mathbf{D}^n_{p}(i,j))},\\
		\mathbf{D}_o &= \sum\nolimits_m\mathbf{V}^m \odot \mathbf{D}_p^m,\\
		\mathbf{A}_o &= \sum\nolimits_m\mathbf{V}^m \odot \mathbf{A}_p^m, \\ 
		\mathbf{p}_o &=\frac{1}{M}\sum\nolimits_m\mathbf{p}_p^m,
	\end{aligned}
\end{equation}
where $V^m_{i,j}$ is the visibility map of the $m$th part capsule at the position $(i,j)$, $\odot$ is the element-wise production, and $\Theta_o:\{\mathbf{D}_o, \mathbf{A}_o, \mathbf{p}_o\}$ is the object capsule. During assembly, a capsule is visible at $(i,j)$ only when its depth is smaller than the others. The part-level depth and albedo maps are multiplied with their visibility maps and aggregated as one, respectively, and the object pose is the average of part poses. In the object capsule, both the depth $\mathbf{D}_o$ and albedo $\mathbf{A}_o$ are defined in the canonical space, and the pose $\mathbf{p}_o$ is used to project the 3D object to the image plane. Finally, by estimating the lighting $\mathbf{l}$ with another module similar to ~\cite{wu2020unsupervised}, the recovered image $\hat{\mathbf{I}}$ is generated by the differentiable rendering $\Lambda$~\cite{kato2018neural}: 
\begin{align}\label{equ-overview-obj}
	\hat{\mathbf{I}} = \Lambda(\mathbf{D}, \mathbf{A}, \mathbf{p}, \mathbf{l}).
\end{align}
When training IGC-Net, we can minimize the distance between the input image $\mathbf{I}$ and the reconstructed image $\hat{\mathbf{I}}$ following the analysis-by-synthesis strategy, so that the network parameters can be learned in an unsupervised manner.

\begin{figure}
	\begin{center}
		\includegraphics[width=1.0\linewidth]{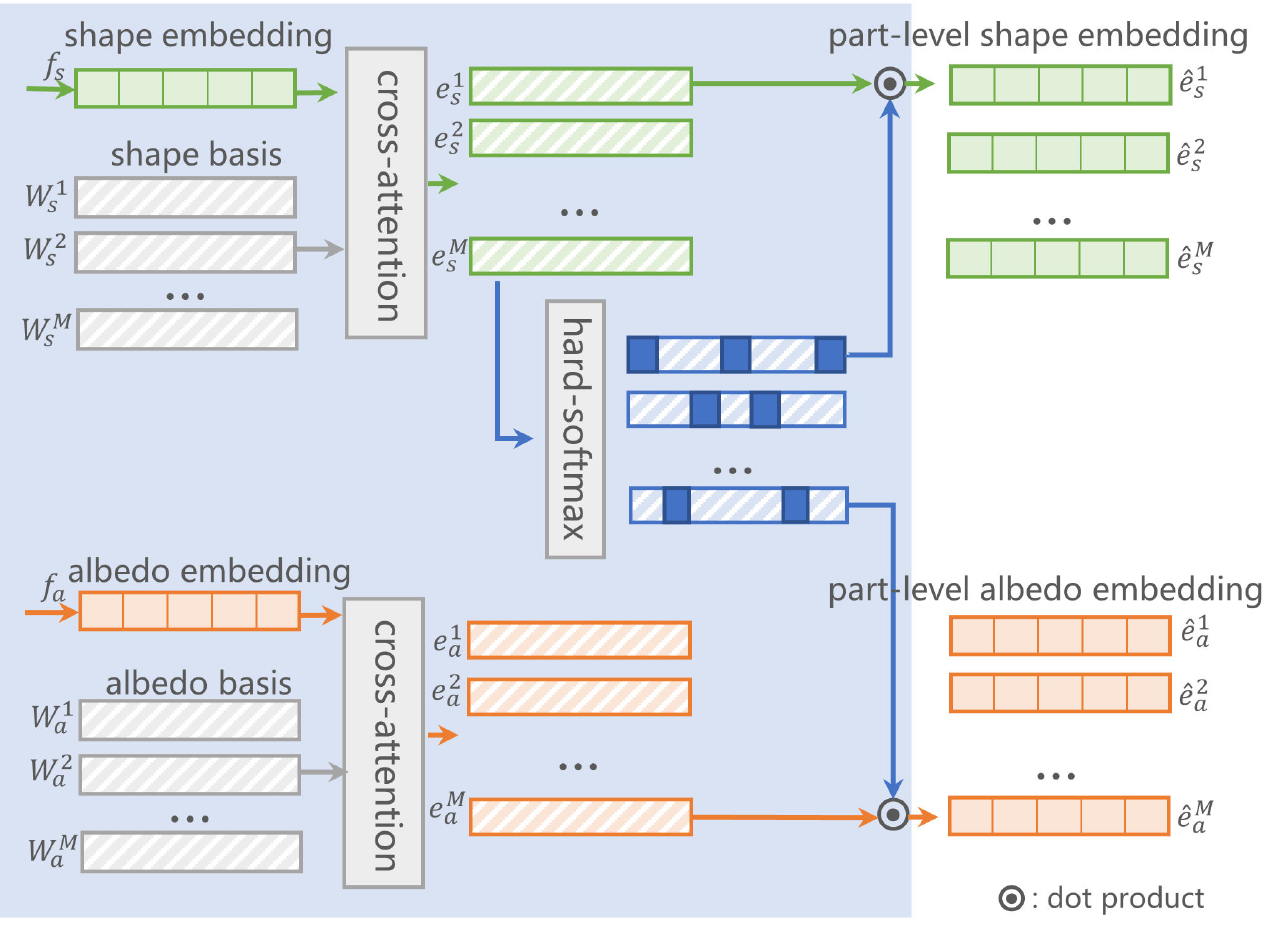}
	\end{center}
	\vspace{-10px}
	\caption{Illustration of the Graphics Decomposition Module (GDM). GDM is proposed to ensure that each part capsule presents a semantic-consistent part of objects.}
	\label{fig-GDM}
\end{figure}

\subsection{Graphics Decomposition Module}
\label{method-attention}
Humans can decompose an object into a set of parts and construct a hierarchy by just observation. To realize this ability in neural networks, we propose the Graphics Decomposition Module (GDM) to decompose the global embedding of the image into a set of semantic-consistent part-level descriptions. The illustration of GDM is shown in Figure~\ref{fig-GDM}.

Taking shape decomposition as an example, GDM maintains $M$ shape basis $\{\mathbf{W}_s^m\}$ as the implicit part templates. Given the global embeddings $\mathbf{f}_s$ extracted in Eqn.~\ref{equ-image-encoder}, GDM performs cross attention between the global embedding and the basis to get $M$ disentangled $D$ dimensional embeddings:
\begin{align}\label{equ-GDM-1}
	\begin{aligned}
		\mathbf{e}_s^m&=\mathbf{f}_s\mathbf{W}_s^m, ~~~ m = 1,...,M.\\
	\end{aligned}
\end{align}
To further reduce the entanglement between $\{\mathbf{e}_s^m\}$ and generate independent part-level embeddings, an $M$-way one-hot attention vector is generated for each of the $D$ dimensions, by deploying that only one embedding can preserve its value and the others are set to $0$ at each dimension. This dimension attention is formulated as: 
\begin{align}\label{equ-GDM-2}
\begin{aligned}
	&\hat{\mathbf{e}}_s^m=\mathbf{e}_s^m \odot \mathbf{M}_{[m,:]},\\
	&\mathbf{M}_{[:,d]} = {\rm hard\_softmax}([\mathbf{e}_s^1(d), \mathbf{e}_s^2(d), ..., \mathbf{e}_s^M(d)]),\\
	&{\rm hard\_softmax}(\mathbf{e})=\frac{\mathbf{e}}{\sum_i\mathbf{e}(i)}\odot ~ {\rm onehot}(\frac{\mathbf{e}}{\sum_i\mathbf{e}(i)}),
\end{aligned}
\end{align}
where $\mathbf{M}_{M \times D}$ is the attention matrix, whose $m$th row is $\mathbf{M}_{[m,:]}$ and $d$th column is $\mathbf{M}_{[:,d]}$, $\mathbf{e}_s^m(d)$ is the $d$th dimension of the embedding $\mathbf{e}_s^m$, ${\rm onehot}(\cdot)$ is the one-hot operation, and $\hat{\mathbf{e}}_s^m$ is the final part-level shape embedding. The same pipeline is applied to the albedo embeddings, where the only difference is that the attention $\mathbf{M}$ is copied from the shape embeddings, which ensures that the shape and the albedo information are decomposed synchronously.

By incorporating both shape and albedo information as cues, GDM successfully decomposes parts from objects under varied poses and appearances, ensuring that each part capsule represents a semantic-consistent part.

\subsection{Loss and Regularization}
\label{method-loss}
When training IGC-Net with unlabelled images, we employ the following constraints to learn the hierarchical 3D representations effectively:

\noindent\textbf{Reconstruction.} We adopt the negative log-likelihood loss~\cite{wu2020unsupervised} to measure the distance between the original image $\mathbf{I}$ and the reconstructed image $\hat{\mathbf{I}}$:
\begin{align}\label{equ-loss-rec}
	\begin{aligned}
		\mathcal{L}_{rec}=-\frac{1}{\vert\Omega\vert}\sum\ln\frac{1}{\sqrt{2}\sigma}\exp-\frac{\sqrt{2}\vert\hat{\mathbf{I}}-\mathbf{I}\vert}{\sigma}\\
		-\frac{1}{\vert\Omega\vert}\sum\ln\frac{1}{\sqrt{2}\sigma}\exp-\frac{\sqrt{2}\vert\hat{\mathbf{I}}_{flip}-\mathbf{I}\vert}{\sigma},
	\end{aligned}
\end{align}
where $\Omega$ is for normalization and $\sigma\in\mathbb{R}^{H\times W}$ is the confidence map estimated by a network to present the symmetric probability of each position in $\mathbf{I}$, $\hat{\mathbf{I}}_{flip}$ is the image reconstructed with the flipped albedo and shape. Following unsup3d~\cite{wu2020unsupervised}, we also incorporate the perceptual loss to improve the reconstruction results:
\begin{align}\label{equ-loss-percep}
	\begin{aligned}
		\mathcal{L}_{per}=-\frac{1}{\vert\Omega^{(k)}\vert}\sum\ln\frac{1}{\sqrt{2}\sigma^{(k)}}\exp-\frac{\sqrt{2}\vert f^{(k)}(\hat{\mathbf{I}})-f^{(k)}(\mathbf{I})\vert}{\sigma^{(k)}}\\
		-\frac{1}{\vert\Omega^{(k)}\vert}\sum\ln\frac{1}{\sqrt{2}\sigma^{(k)}}\exp-\frac{\sqrt{2}\vert f^{(k)}(\hat{\mathbf{I}}_{flip})-f^{(k)}(\mathbf{I})\vert}{\sigma^{(k)}},
	\end{aligned}
\end{align}
where $f^{(k)}(\cdot)$ is the $k$-th layer of a pre-trained image encoder (VGG~\cite{simonyan2014very} in this paper) and $\sigma^{(k)}$ is the corresponding confidence map.

\noindent\textbf{Semantic Consistency.}
In GDM, shape embedding is used as the cue for part decomposition. To improve the semantic consistency across samples, we employ a contrastive loss on the shape embedding $\hat{\mathbf{e}}_s^m$ of each capsule, which is formulated as:
\begin{small}
\begin{align}\label{equ-loss-contrast}
	\begin{aligned}
		&\mathcal{L}_{contra}=-\sum_{b=1}^B\sum_{m=1}^M \log\\
		&\frac{\sum_{i\neq b}\exp(e_s^{m,(b)}\cdot e_s^{m,(i)}/\tau)}{\sum_{i\neq b}\exp(e_s^{m,(b)}\cdot e_s^{m,(i)}/\tau) + \sum_{j\neq m}\sum_{i\neq b}\exp(e_s^{m,(b)}\cdot e_s^{j,(i)}/\tau)},\\
	\end{aligned}
\end{align}
\end{small}
where $B$ is the batch size, $M$ is the number of part capsules, $\hat{\mathbf{e}}_s^{j,(i)}$ is the shape embedding of the $j$th part that belongs to the $i$th sample. $\mathcal{L}_{contra}$ maximizes the shape similarity between the same capsule across the samples and minimizes the similarity across different capsules. $\tau$ is the hyperparameter utilized to control the discrimination across the negative pairs. 

\noindent\textbf{Sparsity.} To prevent the network from collapsing to use one capsule to describe the whole objects, we employ the spatial sparsity constraint on the visible regions $V^m$ of part capsules:
\begin{align}\label{equ-loss-std}
	\begin{aligned}
		\mathcal{L}_{sparse}={\rm std}(\sum_{i,j}\textbf{V}_{i,j}^m),\\
	\end{aligned}
\end{align}
where ${\rm std}(\cdot)$ calculates the standard deviation, $\textbf{V}^m_{i,j}$ is the visibility map of the $m$th capsule at the position $(i,j)$.

\noindent\textbf{Background Separation.}
The prerequisite for unsupervised part discovery is separating foreground and background so that the network can focus on the objects. To achieve that, previous works incorporate salient maps or the ground-truth foreground masks during training. Instead, we use a specific part capsule to model the background. Note that the graphics capsule can recover the 3D information of the objects without any annotation, the foreground map can be easily estimated by setting a threshold to the depth:
\begin{align}\label{equ-loss-bg}
	\begin{aligned}
		\mathcal{L}_{bg}=\Vert\textbf{V}^{bg}-\widetilde{\textbf{V}}\Vert,~~\widetilde{\textbf{V}}=\textbf{1}_{\textbf{D}_{o}<\gamma},\\
	\end{aligned}
\end{align}
where $\textbf{V}^{bg}$ is the visibility map of the part capsule that is used for background estimation, $\widetilde{\textbf{V}}$ is the external region of the object, $\textbf{D}_o$ is the depth of the object, and $\gamma$ is the threshold for locating the external region.

The final loss functions to train IGC-Net are combined as:
\begin{align}\label{equ-loss-all}
	\begin{aligned}
		\mathcal{L}=&\mathcal{L}_{rec}+\lambda_{per}\mathcal{L}_{per}+\lambda_{contra}\mathcal{L}_{contra} \\
		&+\lambda_{sparse}\mathcal{L}_{sparse} +\lambda_{bg}\mathcal{L}_{bg},\\
	\end{aligned}
\end{align}
where $\lambda_{contra}$, $\lambda_{sparse}$ and $\lambda_{bg}$ are the hyper-parameters to balance different loss functions.	

\section{Experiments}

\noindent\textbf{Implement Details.}
The image encoder, the capsule decoder, and the lighting module of IGC-Net are composed of convolutional neural networks. We set the number of the part-level graphics capsules $M=6$, where one of them is used to model the background. Besides, the hyper-parameters for loss combination are set to be $\lambda_{per} = 0.5, \lambda_{contra}=10^{-5}, \lambda_{sparse}=0.1, \lambda_{bg}=0.1$.
For optimization, we use the Adam optimizer~\cite{kingma2014adam} with $10^{-4}$ learning rate to train the networks on a GeForce RTX 3090 for 60 epochs. More training and evaluation details are provided in the supplementary material.

\noindent\textbf{Datasets.}
Following the recent study for the unsupervised face part discovery~\cite{yu2022hp}, we evaluate IGC-Net on BP4D~\cite{zhang2014bp4d} and Multi-PIE~\cite{gross2010multi}. Both of these two datasets are captured in the controlled environment. To further validate the capability of tackling the images under real-world scenarios, we adopt the CelebA~\cite{liu2015deep} for experiments, which contains over 200K in-the-wild images of real human faces. In the experiments, BP4D and CelebA are used to evaluate the unsupervised face segmentation and Multi-PIE is used for the interpretable face analysis.

\subsection{The Discovered Face Hierarchy}
\begin{figure*}	
	\begin{center}
		\includegraphics[width=1.0\linewidth]{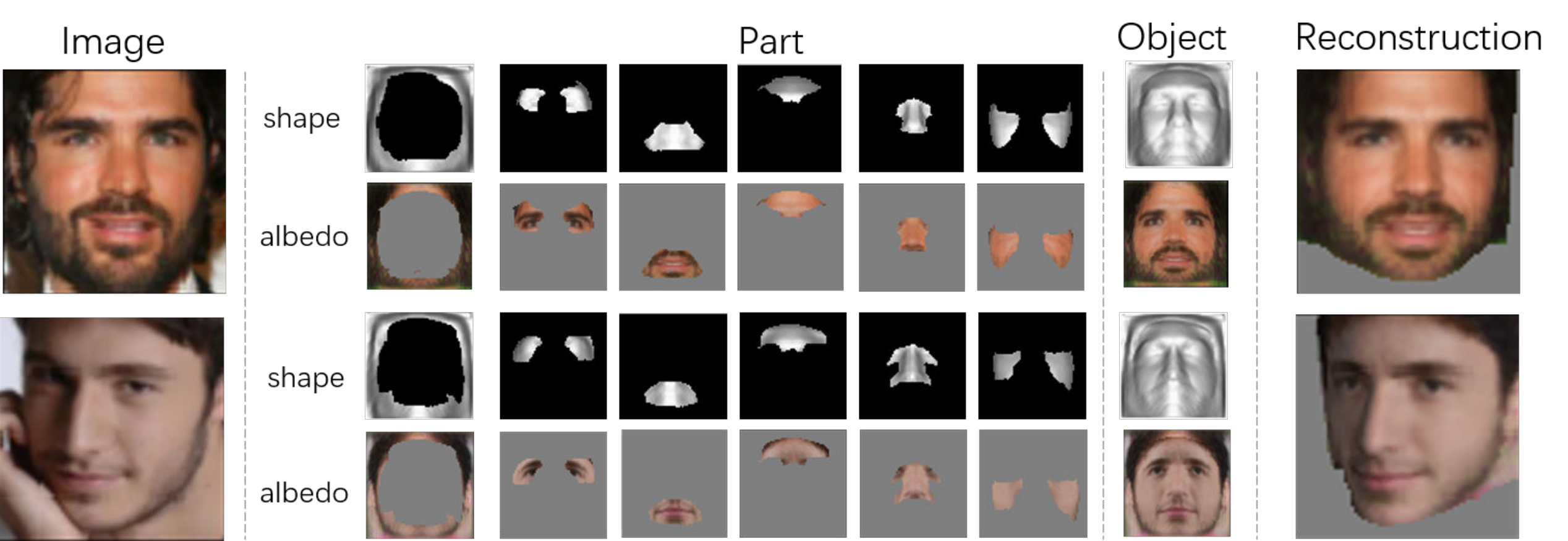}
	\end{center}
	\caption{Illustration of the discovered face hierarchy with 3D descriptions. By incorporating shape and albedo information as cues, IGC-Net decomposes the images into six parts: background, eyes, mouth, forehead, nose, and cheek.}
	\label{fig-hierarchy}
\end{figure*}

Due to the 3D representations embedded in the graphics capsules, IGC-Net successfully builds the hierarchy of in-the-wild faces that are captured under varied illuminations and poses, shown in Figure~\ref{fig-hierarchy}. By incorporating shape and albedo information as cues, the face images are naturally decomposed into six semantic-consistent parts: background, eyes, mouth, forehead, nose, and cheek, without any human supervision. 
Each part is described with a specific graphics capsule, which is composed of a set of interpretable parameters including pose, view-independent shape, and view-independent albedo. These parts are assembled by their depth to generate the object capsules as the object-centered representations, building a bottom-up face hierarchy. We also try to discover other numbers of facial parts by controlling $M$ and get reasonable results, shown in the supplementary material.

\begin{figure}[htbp]
    \centering
    \includegraphics[width=1.\linewidth]{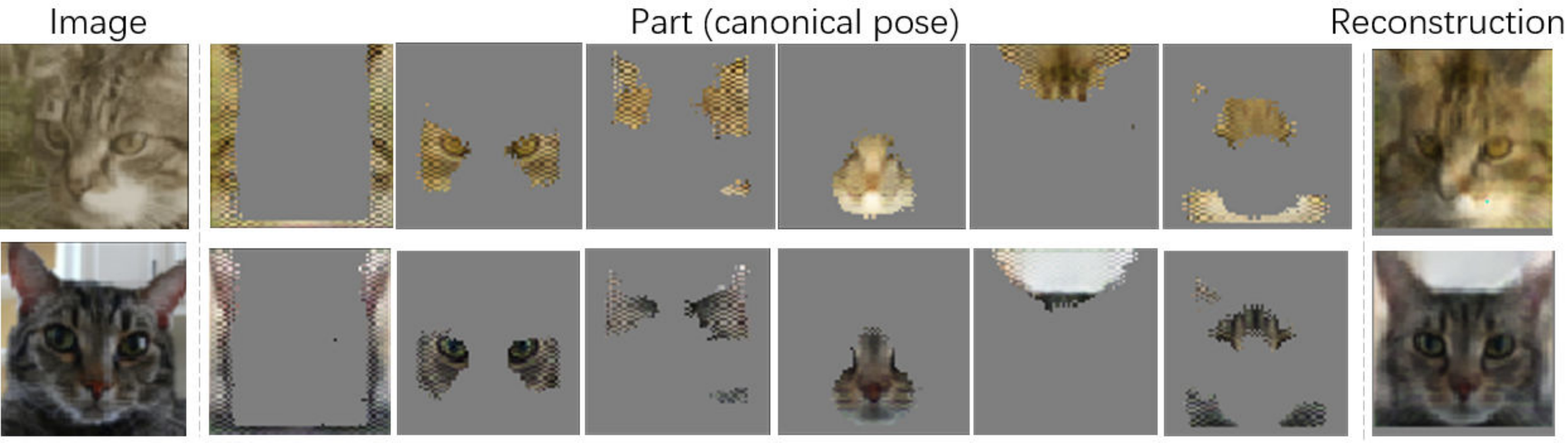}
    \vspace{-4px}
    \caption{The discovered hierarchy of cats. The cat faces are split into background, eyes, ears, nose, forehead, and other skins.}
    \label{fig:cat}
\end{figure}

To show the potential of IGC-Net, we extend our method to image collections of cat faces. Compared with human faces, cat faces are more challenging as cats have more varied textures than humans. The results are shown in Figure~\ref{fig:cat}. It can be seen that the cats are split into background, eyes, ears, nose, forehead, and other skins. 

\subsection{Analysis of Hierarchical 3D Face Representation}
The graphics capsules learned by IGC-Net provide a face hierarchy with explicit graphics descriptions, which gives a plausible way to materialize Marr's theory~\cite{marr1978representation,marr2010vision} that the purpose of vision is building hierarchical 3D representations of objects for recognition. In this section, we apply IGC-Net to validate the advantages of such hierarchical 3D descriptions and uncover the face recognition mechanism of neural networks.

\begin{figure}
\centering
\begin{subfigure}{0.26\linewidth}
	\includegraphics[width=1\linewidth]{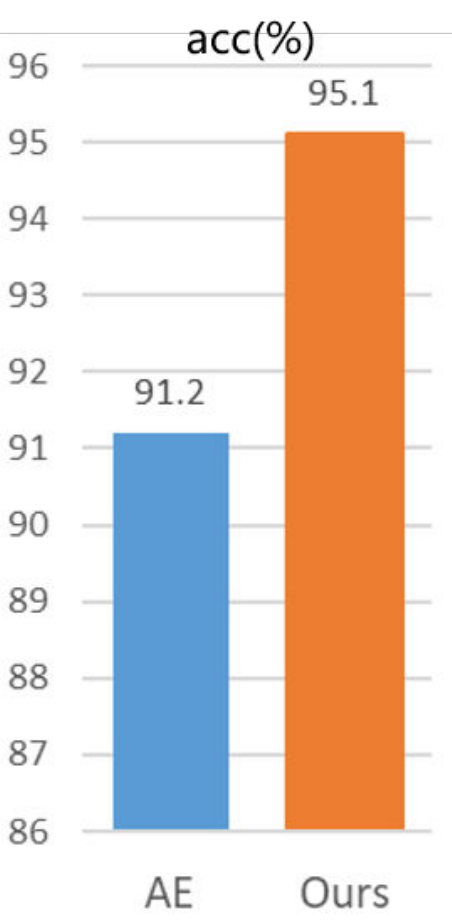}
	\caption{}
	\label{fig-acc-pose}
\end{subfigure}
\hfill
\begin{subfigure}{0.7\linewidth}
	\includegraphics[width=1\linewidth]{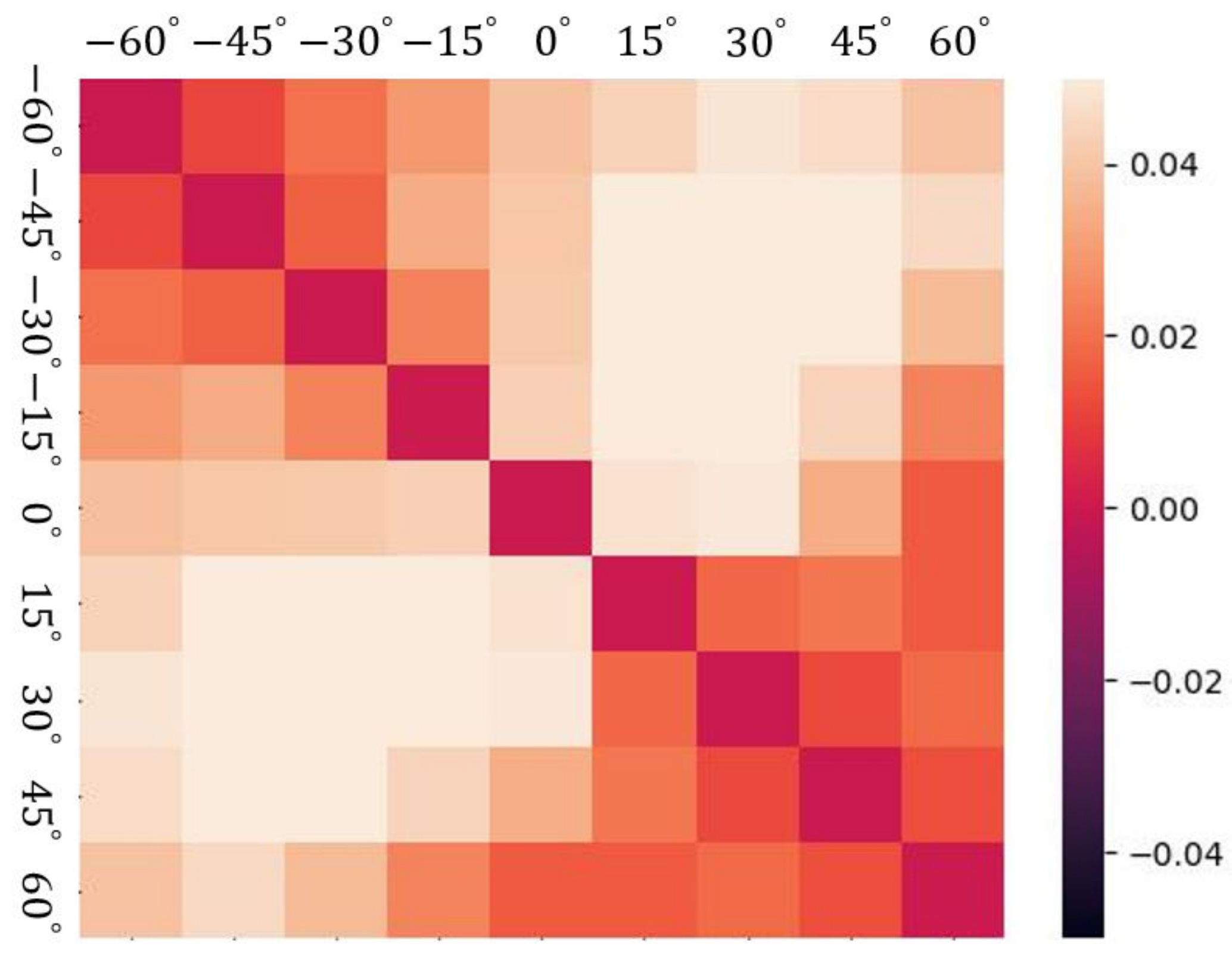}
	\caption{}
	\label{fig-heat-pose}
\end{subfigure}
\caption{The comparison of the representation consistency under different views between 2D representations and 3D representations on Multi-PIE. (a) The recognition accuracy. The representations are sent to a linear classifier for classification. (b) The similarity matrices of the 2D and 3D representations are subtracted and shown as a heatmap. The score higher than 0 indicates 3D is better than 2D.}
\label{fig-pose}
\end{figure}

\noindent\textbf{3D Representation vs. 2D Representation.} 
As Marr's theory reveals~\cite{sutherland1979representation,marr1978representation}, the brain should construct the observer-independent object-centered representations of the objects. To evaluate the view-independence of 2D and 3D representations,  we compare our method with a 2D autoencoder with the architecture and the training strategy same as ours. Specifically, both the models are trained on CelebA and tested on the Multi-PIE dataset with yaw variations from -60 to 60. When performing recognition, the embeddings of the autoencoder, and the depth and albedo embeddings of our method are sent to a linear classifier for face recognition to evaluate the consistency of the representations under different views. The results are shown in Figure~\ref{fig-pose}. Firstly, it can be seen from Figure 4(a) that the 3D presentation achieves better accuracy (95.1\% vs. 91.2\%) in this cross-view recognition task. Secondly, we further analyze the representation consistency across views by computing the similarity matrix of representations under different views. The similarity matrices of the 2D and 3D representations are subtracted (3D minus 2D) and shown as a heatmap in Figure 4(b). It can be seen that our method shows better consistency, especially when matching images across large views, i.e., $30^{\circ}$ vs $-60^{\circ}$.

\begin{figure}[!h]	
	\centering	
	\includegraphics[width=1.0\linewidth]{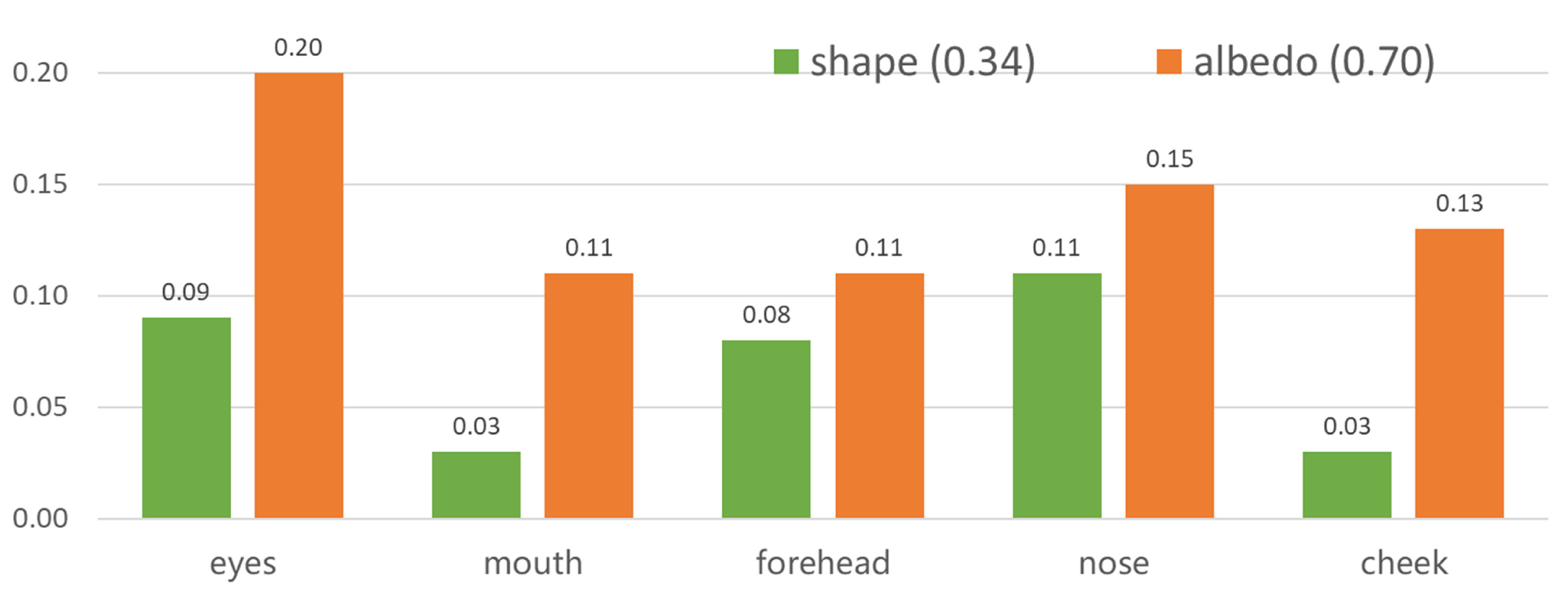}
	\caption{The importance of part-level graphics capsules for face recognition on Multi-PIE. On average, the albedo is more crucial than the shape when recognizing faces. The albedo of the eyes is the most important component and the shape of the nose is more important than the shape of other parts.}
	\label{fig-recog}
\end{figure}

\noindent\textbf{Shape vs. Albedo.} 
To show the potential of our method for interpretable analysis, we design an experiment to explore which part-level graphics capsule is crucial for face recognition and which component (shape or albedo) of the capsule is more important. Specifically, we assign the part-level shape embeddings $\{\hat{\textbf{e}}_s^m\}$ and albedo embeddings $\{\hat{\textbf{e}}_a^m\}$ with trainable scalar $\{\omega_s^m\}$ and $\{\omega_a^m\}$ as the attention weights. The weight parameters $\{\omega^m\hat{\textbf{e}}^m\}$ are sent to a linear classifier for face recognition. After training with L1 penalization for sparsity, the attention weights of part capsules are shown in Figure~\ref{fig-recog}. By summarizing the attention weights of different parts, we can see that the albedo ($\omega=0.70$) is more important than the shape ($\omega=0.34$) for face recognition. Besides, the part-level attention weights also show that the albedo of the eyes is the most important component and the shape of the nose is more important than the shape of other parts, which is consistent with the previous conclusions~\cite{williford2020explainable,yu2022hp}.

\subsection{Unsupervised Face Segmentation}
To execute the quantitative and qualitative evaluation, we treat the silhouettes of parts as segment maps and apply them to the unsupervised face segmentation task. Note that there is no ground truth for the unsupervised part-level segmentation, the key of this task is to evaluate the semantic consistency of the parsing manners. The following experiments show the superiority of our method.

\noindent\textbf{Baselines.}
Learning to segment the face parts from the unlabeled images is a challenging task as parts are difficult to be described by math. In this paper, we compare our method with the state-of-art methods for unsupervised face segmentation, including DFF~\cite{collins2018deep}, SCOPS~\cite{hung2019scops} and HP-Capsule~\cite{yu2022hp}.
To discover the semantic parts, DFF proposes to execute the non-negative matrix upon the CNN features, which need to optimize the whole dataset to get the segment results. SCOPS proposes a framework with the concentration loss to constrain the invariance of images between TPS transformation. However, due to the lack of effective constraints, their results tend to assign similar silhouettes to different parts. HP-Capsule proposes a bottom-up schedule to aggregate parts from subparts, whose parts are described with interpretable parameters. However, their descriptions are defined in the 2D space, limiting their capacity to tackle faces with large poses. 

\begin{figure}		
	\includegraphics[width=1.0\linewidth]{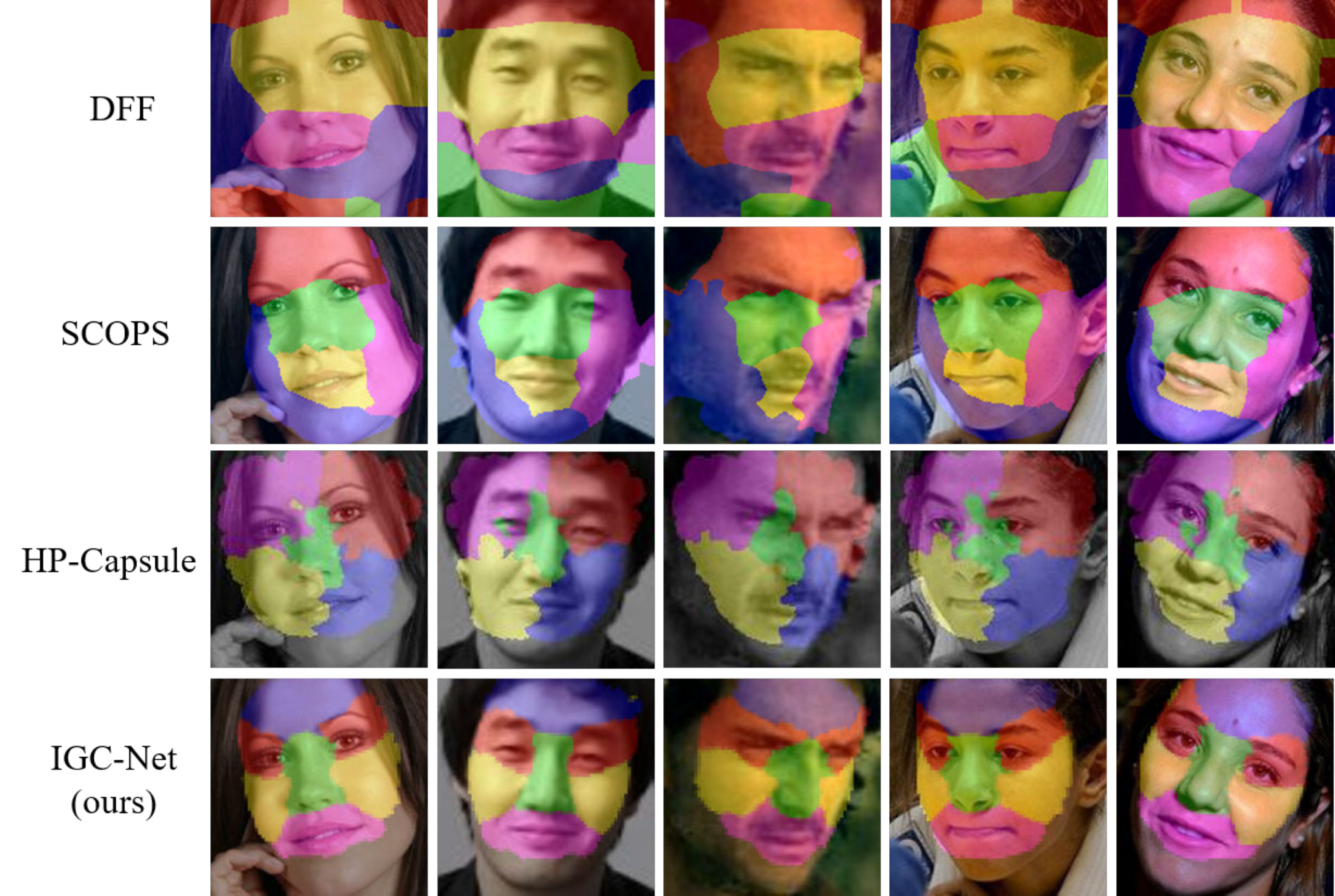}
	\caption{The qualitative comparison of unsupervised face segmentation on CelebA.}
	\label{fig-seg-celebA}						
\end{figure}

\begin{figure}		
	\includegraphics[width=1.0\linewidth]{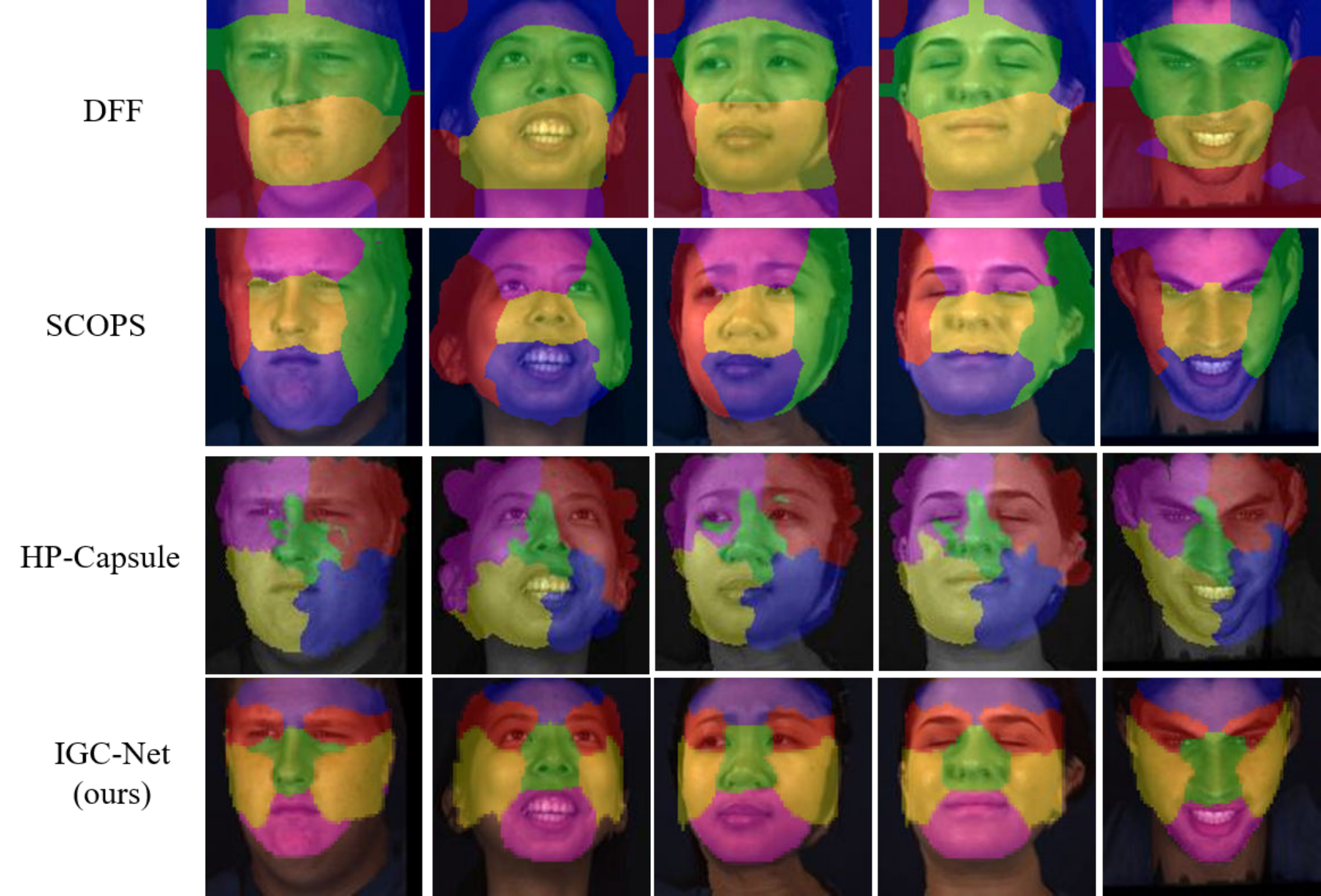}
	\caption{The qualitative comparison of unsupervised face segmentation on BP4D.}
	\label{fig-seg-BP4D}						
\end{figure}

\begin{table}	
	\centering
	\caption{The quantitative comparison of unsupervised face segmentation on CelebA. $\rm NME_L (\%)$ and $\rm NME_{DL} (\%)$ use the landmarks estimated from the segment maps to evaluate the semantic consistency of parts.}
	\label{tab-seg-celeba}
	\setlength{\tabcolsep}{4mm}
	\begin{tabular}{lll}
	\toprule
		\multicolumn{1}{c}{METHOD}  &\multicolumn{1}{c}{ $\rm{NME_L}$} & \multicolumn{1}{c}{\bf $\rm{NME_{DL}}$}
		\\ \midrule		
		DFF~\cite{collins2018deep}  & 22.78   & 27.27  \\
		SCOPS~\cite{hung2019scops}   & 18.72  & 23.69 \\
		HP-Capsule~\cite{yu2022hp} & 21.25  & 25.27  \\
		IGC-Net (ours) &\textbf{11.84}   & \textbf{18.88}   \\
	\bottomrule
	\end{tabular}
\end{table}

\begin{table}		
	\centering
	\caption{The quantitative comparison of unsupervised face segmentation on BP4D.}
	\setlength{\tabcolsep}{4mm}
	\label{tab-seg-BP4D}
	\begin{tabular}{lll}
	\toprule
		\multicolumn{1}{c}{METHOD}  &\multicolumn{1}{c}{ $\rm{NME_L}$} & \multicolumn{1}{c}{\bf $\rm{NME_{DL}}$}
		\\ \midrule
		DFF~\cite{collins2018deep}  & 18.85   & 12.26  \\
		SCOPS~\cite{hung2019scops}   & 9.10  &  6.74 \\
		HP-Capsule~\cite{yu2022hp} &8.81   & 6.10   \\
		IGC-Net (ours) &\textbf{6.35}   & \textbf{4.32}   \\
	\bottomrule
	\end{tabular}
	
\end{table}

\begin{figure}	
	\includegraphics[width=1.0\linewidth]{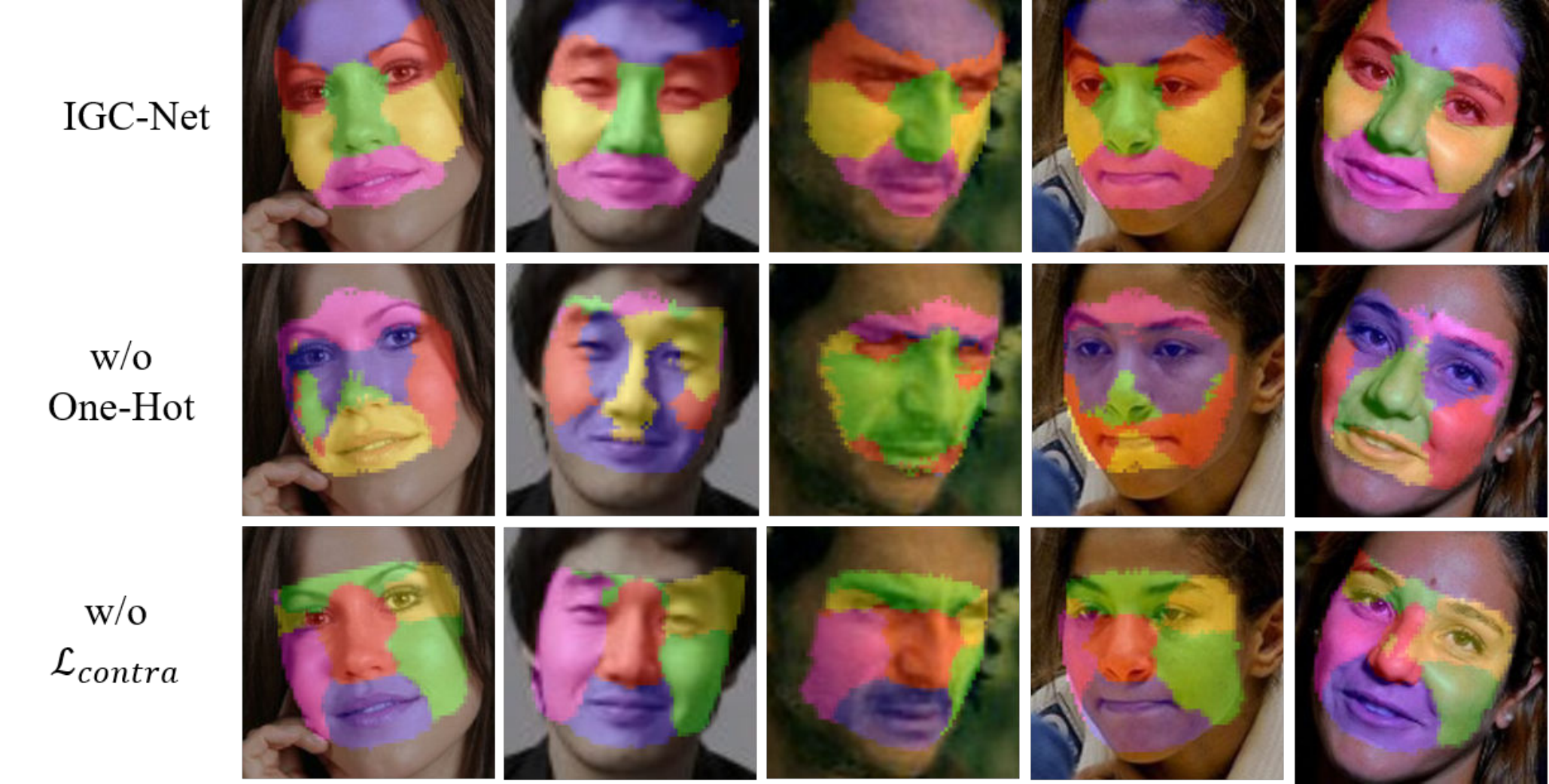}
	\caption{The qualitative ablation study on CelebA. It can be seen that the semantic consistency will be damaged without the one-hot operation in GDM (see Eqn.~\ref{equ-GDM-2}) and the $\mathcal{L}_{contra}$ (see Eqn.~\ref{equ-loss-contrast}) is important for discovering parts with salient semantics.}
	\label{fig-ablation}						
\end{figure}


\noindent\textbf{Quantitative Comparison}. Following the previous work~\cite{yu2022hp}, we utilize the Normalized Mean Error (NME) of the landmarks predicted by segment maps to evaluate the quality of the parsing manners. Specifically, $\rm NME_L$ treats the centroid of the segment maps as landmarks and uses linear mapping to convert them to human-annotated landmarks. $\rm NME_{DL}$ incorporates a shallow network to directly predict the landmarks from the segment maps. Table~\ref{tab-seg-celeba} and Table~\ref{tab-seg-BP4D} show the quantitative comparison results on CelebA and BP4D, which validate the effectiveness of our method.

\noindent\textbf{Qualitative Comparison}. The qualitative comparison results are shown in Figure~\ref{fig-seg-celebA} and Figure~\ref{fig-seg-BP4D}. It can be seen that our method performs better than other methods. The results of DFF don't successfully separate the foreground and the background. As for SCOPS, due to the lack of effective constraints, the segment maps of SCOPS are with some ambiguity, where the organs with salient semantics are assigned to different parts for different samples. For example, SCOPS sometimes takes the right eye as the green part (the fifth column in Figure~\ref{fig-seg-celebA}) while sometimes splitting it from the middle (the first and the second column in Figure~\ref{fig-seg-celebA}). The segment boundaries of HP-Capsule are clearer than DFF and SCOPS. However, as shown in the third column of Figure~\ref{fig-seg-celebA}, limited by their 2D descriptions, HP-Capsule fails on the faces with large poses while our method performs well on these challenging samples.

\begin{table}	
	\centering
	\caption{The quantitative ablation study on CelebA. The results show the importance of the one-hot operation in GDM and the semantic constraint $\mathcal{L}_{contra}$.}
	\setlength{\tabcolsep}{4mm}
	\label{tab-ablation}
	\begin{tabular}{ll|l}
		\toprule
		One-Hot & $\mathcal{L}_{contra}$ & $\rm{NME_{L}}$\\ \midrule
		&$\checkmark$& 19.10 \\
		$\checkmark$  &  & 13.46\\
		$\checkmark$  & $\checkmark$  & \textbf{11.84}  \\
		\bottomrule
	\end{tabular}			
\end{table}

\subsection{Ablation Studies}
The basis of building the hierarchy of objects is to learn the parts with explicit semantics and keep semantic consistency across different samples. In this section, we perform the ablation study to show the importance of the one-hot operation in the GDM (see Eqn.~\ref{equ-GDM-2}) and the semantic constraint $\mathcal{L}_{contra}$ (see Eqn.~\ref{equ-loss-contrast}) for discovering meaningful parts. Figure~\ref{fig-ablation} shows the qualitative ablation study on CelebA. In the second row of Figure~\ref{fig-ablation}, it can be seen that, without the one-hot operation to prevent the information leakage of different parts, the semantic consistency across samples will be damaged. The third row of Figure~\ref{fig-ablation} shows that the contrastive semantic constraint $\mathcal{L}_{contra}$ is important for the discovery of parts with salient semantics. Without such constraint, the segmentation of the important organs such as the eyes will have ambiguity. These conclusions are also validated by the quantitative ablation study shown in Table~\ref{tab-ablation}.

\section{Conclusion and Discussion}
In this paper, we propose the IGC-Net to learn the hierarchical 3D face representations from large-scale unlabeled in-the-wild images, whose primitive is the graphics capsule that contains the 3D representations with explicit meanings. By combining depth and albedo information as cues, IGC-Net successfully decomposes the objects into a set of part-level graphics capsules and constructs the hierarchy of objects by assembling the part-level capsules into object-level capsules. IGC-Net reveals how the neural networks, oriented at visual perception, understand faces as a hierarchy of 3D models. Besides, the part-level graphics descriptions can be used for unsupervised face segmentation and interpretable face analysis. Experiments on CelebA, BP4D, and Multi-PIE validate the effectiveness and the interpretability of our method.

\noindent\textbf{Acknowledgement.} This work was supported in part by the National Key Research \& Development Program (No. 2020YFC2003901), Chinese National Natural Science Foundation Projects \#62276254, \#62176256, \#62206280, the Youth Innovation Promotion Association CAS (\#Y2021131), and the InnoHK program.

{\small
\bibliographystyle{ieee_fullname}
\bibliography{egbib}
}

\end{document}